\documentclass[conference]{IEEEtran}
\usepackage{cite}
\usepackage{graphicx}
\usepackage{float}
\usepackage{amsmath}
\usepackage{listings}
\usepackage{xcolor}
\begin{document}

\title{Computer Vision-based Accident Detection in Traffic Surveillance}
\author
{
\IEEEauthorblockN{Earnest Paul Ijjina\IEEEauthorrefmark{1}}
\IEEEauthorblockA{Assistant Professor, Department of Computer Science and Engineering\\
National Institute of Technology Warangal, India-506004\\
Email : \IEEEauthorrefmark{1}  iep@nitw.ac.in}

\IEEEauthorblockN{Dhananjai Chand\IEEEauthorrefmark{2}, Savyasachi Gupta \IEEEauthorrefmark{3}, Goutham K \IEEEauthorrefmark{4}}
\IEEEauthorblockA{B.Tech., Department of Computer Science and Engineering\\
National Institute of Technology Warangal, India-506004\\
Email : \IEEEauthorrefmark{2} cdhananjai@student.nitw.ac.in, \IEEEauthorrefmark{3} gsavyasachi@student.nitw.ac.in, \IEEEauthorrefmark{4}kgoutham@student.nitw.ac.in}
}
\maketitle

\begin{abstract}

Computer vision-based accident detection through video surveillance has become a beneficial but daunting task. In this paper, a neoteric framework for detection of road accidents is proposed. The proposed framework capitalizes on Mask R-CNN for accurate object detection followed by an efficient centroid based object tracking algorithm for surveillance footage. The probability of an accident is determined based on speed and trajectory anomalies in a vehicle after an overlap with other vehicles. The proposed framework provides a robust method to achieve a high Detection Rate and a low False Alarm Rate on general road-traffic CCTV surveillance footage. This framework was evaluated on diverse conditions such as broad daylight, low visibility, rain, hail, and snow using the proposed dataset. This framework was found effective and paves the way to the development of general-purpose vehicular accident detection algorithms in real-time.

\end{abstract}
\begin{IEEEkeywords}Accident Detection, Mask R-CNN, Vehicular Collision, Centroid based Object Tracking
\end{IEEEkeywords}

\section{Introduction}

Vehicular Traffic has become a substratal part of people's lives today and it affects numerous human activities and services on a diurnal basis. Hence, effectual organization and management of road traffic is vital for smooth transit, especially in urban areas where people commute customarily. Annually, human casualties and damage of property is skyrocketing in proportion to the number of vehicular collisions and production of vehicles \cite{ref2}. Despite the numerous measures being taken to upsurge road monitoring technologies such as CCTV cameras at the intersection of roads \cite{ref3} and radars commonly placed on highways that capture the instances of over-speeding cars \cite{ref4, ref31, ref32} , many lives are lost due to lack of timely accidental reports \cite{ref2} which results in delayed medical assistance given to the victims. Current traffic management technologies heavily rely on human perception of the footage that was captured. This takes a substantial amount of effort from the point of view of the human operators and does not support any real-time feedback to spontaneous events.\par
Statistically, nearly 1.25 million people forego their lives in road accidents on an annual basis with an additional 20-50 million injured or disabled. Road traffic crashes ranked as the 9th leading cause of human loss and account for 2.2 per cent of all casualties worldwide \cite{ref10}. They are also predicted to be the fifth leading cause of human casualties by 2030 \cite{ref10}.\par
In recent times, vehicular accident detection has become a prevalent field for utilizing computer vision \cite{ref11} to overcome this arduous task of providing first-aid services on time without the need of a human operator for monitoring such event. Hence, this paper proposes a pragmatic solution for addressing aforementioned problem by suggesting a solution to detect Vehicular Collisions almost spontaneously which is vital for the local paramedics and traffic departments to alleviate the situation in time. This paper introduces a solution which uses state-of-the-art supervised deep learning framework \cite{ref1} to detect many of the well-identified road-side objects trained on well developed training sets\cite{ref12}. We then utilize the output of the neural network to identify road-side vehicular accidents by extracting feature points and creating our own set of parameters which are then used to identify vehicular accidents. This method ensures that our approach is suitable for real-time accident conditions which may include daylight variations, weather changes and so on. Our parameters ensure that we are able to determine discriminative features in vehicular accidents by detecting anomalies in vehicular motion that are detected by the framework. Additionally, we plan to aid the human operators in reviewing past surveillance footages and identifying accidents by being able to recognize vehicular accidents with the help of our approach.\par
The layout of the rest of the paper is as follows. Section \ref{section1} succinctly debriefs related works and literature. Section \ref{section2} delineates the proposed framework of the paper. Section \ref{section3} contains the analysis of our experimental results. Section \ref{section4} illustrates the conclusions of the experiment and discusses future areas of exploration. 

\section{Related Work}
\label{section1}
Over a course of the precedent couple of decades, researchers in the fields of image processing and computer vision have been looking at traffic accident detection with great interest \cite{ref11}. As a result, numerous approaches have been proposed and developed to solve this problem.\par

One of the solutions,  proposed by Singh \emph{et al.} to detect vehicular accidents used the feed of a CCTV surveillance camera by generating Spatio-Temporal Video Volumes (STVVs) and then extracting deep representations on denoising autoencoders in order to generate an anomaly score while simultaneously detecting moving objects, tracking the objects, and then finding the intersection of their tracks to finally determine the odds of an accident occurring. This approach may effectively determine car accidents in intersections with normal traffic flow and good lighting conditions. However, it suffers a major drawback in accurate predictions when determining accidents in low-visibility conditions, significant occlusions in car accidents, and large variations in traffic patterns \cite{ref13}. Additionally, it performs unsatisfactorily because it relies only on trajectory intersections and anomalies in the traffic flow pattern, which indicates that it won’t perform well in erratic traffic patterns and non-linear trajectories.\par

Similarly, Hui \emph{et al.} suggested an approach which uses the Gaussian Mixture Model (GMM) to detect vehicles and then the detected vehicles are tracked using the mean shift algorithm. Even though this algorithm fairs quite well for handling occlusions during accidents, this approach suffers a major drawback due to its reliance on limited parameters in cases where there are erratic changes in traffic pattern and severe weather conditions \cite{ref14}.\par
Moreover, Ki \emph{et al.} have demonstrated an approach that has been divided into two parts. The first part takes the input and uses a form of gray-scale image subtraction to detect and track vehicles. The second part applies feature extraction to determine the tracked vehicle’s acceleration, position, area, and direction. The approach determines the anomalies in each of these parameters and based on the combined result, determines whether or not an accident has occurred based on pre-defined thresholds \cite{ref15}. Even though their second part is a robust way of ensuring correct accident detections, their first part of the method faces severe challenges in accurate vehicular detections such as, in the case of environmental objects obstructing parts of the screen of the camera, or similar objects overlapping their shadows and so on. \par

Though these given approaches keep an accurate track of motion of the vehicles but perform poorly in parametrizing the criteria for accident detection. They do not perform well in establishing standards for accident detection as they require specific forms of input and thereby cannot be implemented for a general scenario.
The existing approaches are optimized for a single CCTV camera through parameter customization. However, the novelty of the proposed framework is in its ability to work with any CCTV camera footage.

\section{Proposed Approach}
\label{section2}
This section describes our proposed framework given in Figure \ref{fig:workflow}. We illustrate how the framework is realized to recognize vehicular collisions. Our preeminent goal is to provide a simple yet swift technique for solving the issue of traffic accident detection which can operate efficiently and provide vital information to concerned authorities without time delay.\\
The proposed accident detection algorithm includes the following key tasks:
\begin{enumerate}
\item [T1:] Vehicle Detection
\item [T2:] Vehicle Tracking and Feature Extraction
\item [T3:] Accident Detection
\end{enumerate}
The proposed framework realizes its intended purpose via the following stages: \\

\subsection{Vehicle Detection}
\label{subsec1}
This phase of the framework detects vehicles in the video. \\

\begin{figure}[h] 
\centering
\includegraphics[width=70mm, height=80mm]{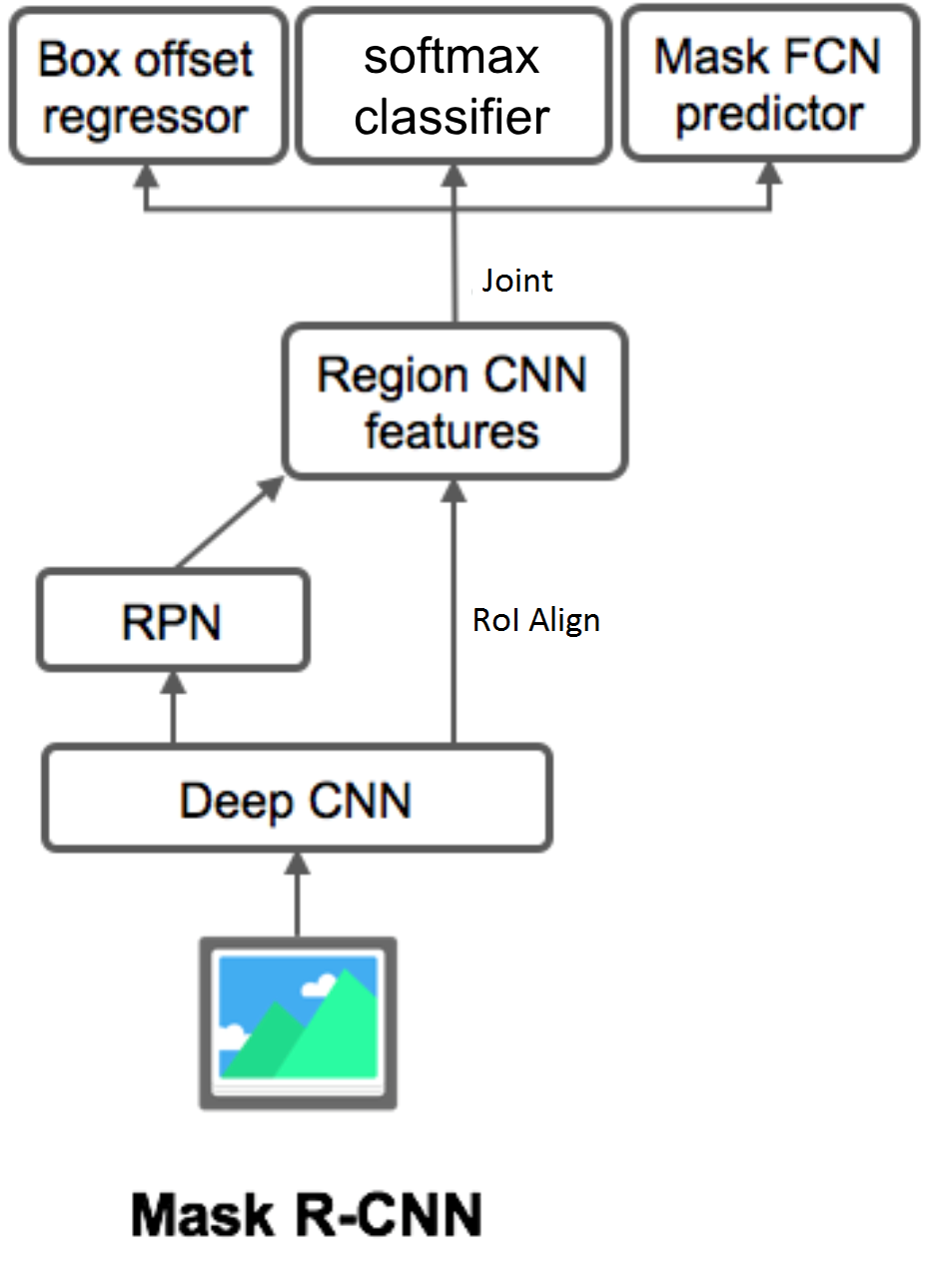}
\caption{The Mask R-CNN framework (from \cite{ref16})}
\label{fig:framework}
\end{figure}

The object detection framework used here is Mask R-CNN (Region-based Convolutional Neural Networks) as seen in Figure \ref{fig:framework}. Using Mask R-CNN we automatically segment and construct pixel-wise masks for every object in the video. Mask R-CNN is an instance segmentation algorithm that was introduced by He \emph{et al.} \cite{ref1}. Mask R-CNN improves upon Faster R-CNN \cite{ref6} by using a new methodology named as RoI Align instead of using the existing RoI Pooling which provides 10\% to 50\% more accurate results for masks\cite{ref1}. This is achieved with the help of RoI Align by overcoming the location misalignment issue suffered by RoI Pooling which attempts to fit the blocks of the input feature map. Mask R-CNN not only provides the advantages of Instance Segmentation but also improves the core accuracy by using RoI Align algorithm.  The result of this phase is an output dictionary containing all the class IDs, detection scores, bounding boxes, and the generated masks for a given video frame.\par

\subsection{Vehicle Tracking and Feature Extraction}
\label{subsec2}

After the object detection phase, we filter out all the detected objects and only retain correctly detected vehicles on the basis of their class IDs and scores. Once the vehicles have been detected in a given frame, the next imperative task of the framework is to keep track of each of the detected objects in subsequent time frames of the footage. This is accomplished by utilizing a simple yet highly efficient object tracking algorithm known as Centroid Tracking \cite{ref34}. This algorithm relies on taking the Euclidean distance between centroids of detected vehicles over consecutive frames. From this point onwards, we will refer to vehicles and objects interchangeably.\par

The centroid tracking mechanism used in this framework is a multi-step process which fulfills the aforementioned requirements. The following are the steps:
\begin{enumerate}
\item The centroid of the objects are determined by taking the intersection of the lines passing through the mid points of the boundary boxes of the detected vehicles.
\item Calculate the Euclidean distance between the centroids of newly detected objects and existing objects.
\item Update coordinates of existing objects based on the shortest Euclidean distance from the current set of centroids and the previously stored centroid.
\item Register new objects in the field of view by assigning a new unique ID and storing its centroid coordinates in a dictionary.
\item De-register objects which haven't been visible in the current field of view for a predefined number of frames in succession.

\begin{figure*}
\includegraphics[width=\textwidth, height = 170mm]{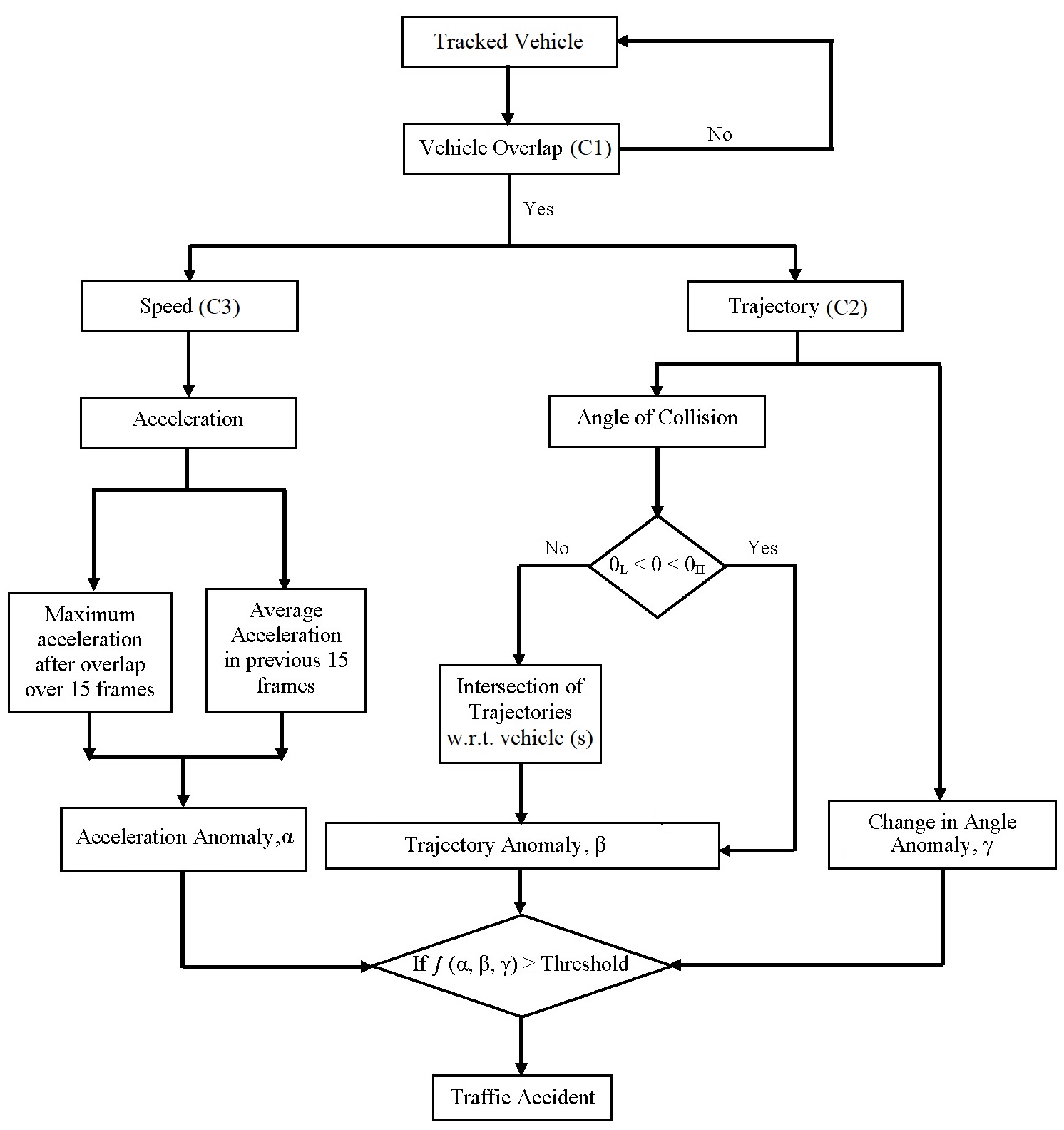}
\caption{Workflow diagram describing the process of accident detection. }
\label{fig:workflow}
\end{figure*}

\end{enumerate}

The primary assumption of the centroid tracking algorithm used is that although the object will move between subsequent frames of the footage, the distance between the centroid of the same object between two successive frames will be less than the distance to the centroid of any other object. This explains the concept behind the working of Step 3.\par

Once the vehicles are assigned an individual centroid, the following criteria are used to predict the occurrence of a collision as depicted in Figure \ref{fig:workflow}.
\begin{enumerate}
\item [C1:] The overlap of bounding boxes of vehicles
\item [C2:] Determining Trajectory and their angle of intersection
\item [C3:] Determining Speed and their change in acceleration
\end{enumerate}

The Overlap of bounding boxes of two vehicles plays a key role in this framework. Before the collision of two vehicular objects, there is a high probability that the bounding boxes of the two objects obtained from Section \ref{subsec1} will overlap. However, there can be several cases in which the bounding boxes do overlap but the scenario does not necessarily lead to an accident. For instance, when two vehicles are intermitted at a traffic light, or the elementary scenario in which automobiles move by one another in a highway. This could raise false alarms, that is why the framework utilizes other criteria in addition to assigning nominal weights to the individual criteria. 

The process used to determine, where the bounding boxes of two vehicles overlap goes as follow:\\
Consider a, b to be the bounding boxes of two vehicles A and B. Let x, y be the coordinates of the centroid of a given vehicle and let $\alpha$, $\beta$ be the width and height of the bounding box of a vehicle respectively. At any given instance, the bounding boxes of A and B overlap, if the condition shown in Eq. \ref{eq:1} holds true.
\begin{equation}\label{eq:1}
( 2 \times |a.x - b.x| < a.\alpha + b.\alpha) \land ( 2 \times |a.y + b.y| < a.\beta + b.\beta)
\end{equation}

The condition stated above checks to see if the centers of the two bounding boxes of A and B are close enough that they will intersect. This is done for both the axes. If the boxes intersect on both the horizontal and vertical axes, then the boundary boxes are denoted as intersecting. This is a cardinal step in the framework and it also acts as a basis for the other criteria as mentioned earlier.\par

The next task in the framework, T2, is to determine the trajectories of the vehicles. This is determined by taking the differences between the centroids of a tracked vehicle for every five successive frames which is made possible by storing the centroid of each vehicle in every frame till the vehicle's centroid is registered as per the centroid tracking algorithm mentioned previously. This results in a 2D vector, representative of the direction of the vehicle’s motion. We then determine the magnitude of the vector, $\boldsymbol{\mu}$, as shown in Eq. \ref{eq:2}.
\begin{equation}\label{eq:2}
\text{magnitude}= \sqrt {\left( {\boldsymbol{\mu}.i } \right)^2 + \left( {\boldsymbol{\mu}.j } \right)^2 }
\end{equation}

We then normalize this vector by using scalar division of the obtained vector by its magnitude. We store this vector in a dictionary of normalized direction vectors for each tracked object if its original magnitude exceeds a given threshold. Otherwise, we discard it. This is done in order to ensure that minor variations in centroids for static objects do not result in false trajectories. We then display this vector as trajectory for a given vehicle by extrapolating it.\par

Then, we determine the angle between trajectories by using the traditional formula for finding the angle between the two direction vectors. Here, we consider $\boldsymbol{\mu_1}$ and $\boldsymbol{\mu_2}$ to be the direction vectors for each of the overlapping vehicles respectively. Then, the angle of intersection between the two trajectories $\theta$ is found using the formula in Eq. \ref{eq:3}.
\begin{equation}\label{eq:3}
\theta = \arccos \left(\frac{\boldsymbol{\mu_1 \cdot \mu_2}}{\boldsymbol{|\mu_1| |\mu_2|}}\right)
\end{equation}
We will discuss the use of $\theta$ and introduce a new parameter to describe the individual occlusions of a vehicle after a collision in Section \ref{subsec3}.\par

The next criterion in the framework, C3, is to determine the speed of the vehicles. We determine the speed of the vehicle in a series of steps. We estimate $\tau$, the interval between the frames of the video, using the Frames Per Second (FPS) as given in Eq. \ref{eq:4}.
\begin{equation}\label{eq:4}
\tau = \frac{1}{\text{FPS}}
\end{equation}

Then, we determine the distance covered by a vehicle over five frames from the centroid of the vehicle ${c_1}$ in the first frame and ${c_2}$ in the fifth frame. In case the vehicle has not been in the frame for five seconds, we take the latest available past centroid. We then determine the Gross Speed (${S_g}$) from centroid difference taken over the ${Interval}$ of five frames using Eq. \ref{eq:5}.
\begin{equation}\label{eq:5}
S_g = \frac{c_2 - c_1}{\tau \times Interval}
\end{equation}

Next, we normalize the speed of the vehicle irrespective of its distance from the camera using Eq. \ref{eq:6} by taking the height of the video frame (${H}$) and the height of the bounding box of the car (${h}$) to get the Scaled Speed (${S_s}$) of the vehicle. The Scaled Speeds of the tracked vehicles are stored in a dictionary for each frame. 
\begin{equation}\label{eq:6}
S_s = \left(\frac{H - h}{H} + 1\right) \times S_g
\end{equation}

Then, the Acceleration (${A}$) of the vehicle for a given ${Interval}$ is computed from its change in Scaled Speed from ${S_s^1}$ to ${S_s^2}$ using Eq. \ref{eq:7}.
\begin{equation}\label{eq:7}
A = \frac{S_s^2 - S_s^1}{\tau \times Interval}
\end{equation}

The use of change in Acceleration (${A}$) to determine vehicle collision is discussed in Section \ref{subsec3}.\par

\begin{figure}[H] 
\centering
\includegraphics[width=85mm, height=80mm]{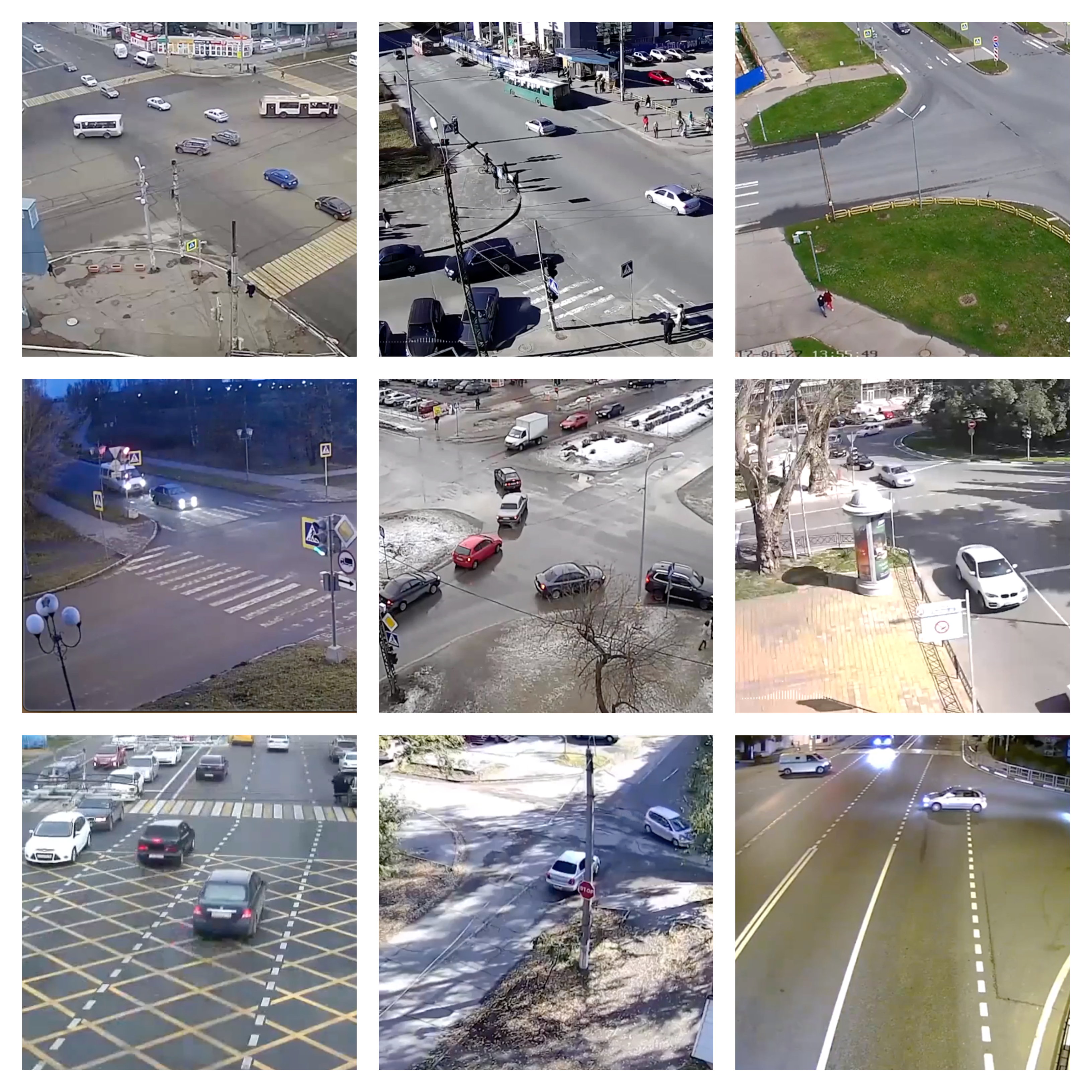}
\caption{Videos depicting various traffic and environmental conditions in the dataset used to evaluate the performance of the proposed framework.}
\label{fig:conditions}
\end{figure}

\subsection{Accident Detection}
\label{subsec3}

This section describes the process of accident detection when the vehicle overlapping criteria (C1, discussed in Section \ref{subsec2}) has been met as shown in Figure \ref{fig:workflow}. We will introduce three new parameters (${\alpha, \beta, \gamma}$) to monitor anomalies for accident detections. The parameters are:
\begin{enumerate}
\item Acceleration Anomaly, $\alpha$
\item Trajectory Anomaly, $\beta$
\item Change in Angle Anomaly, $\gamma$
\end{enumerate}

\begin{figure*}
\includegraphics[width=\textwidth, height = 30mm]{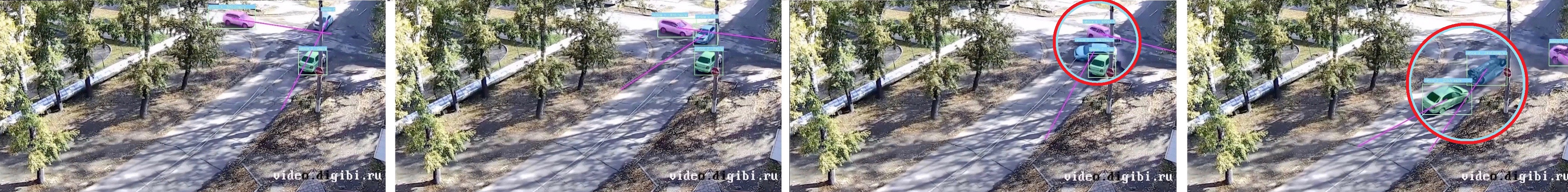}
\caption{An illustration depicting the sequence of frames that lead to the detection of a vehicular accident.}
\label{fig:accidents}
\end{figure*}

When two vehicles are overlapping, we find the acceleration of the vehicles from their speeds captured in the dictionary. We find the average acceleration of the vehicles for 15 frames before the overlapping condition (C1) and the maximum acceleration of the vehicles 15 frames after C1. We find the change in accelerations of the individual vehicles by taking the difference of the maximum acceleration and average acceleration during overlapping condition (C1). The Acceleration Anomaly ($\alpha$) is defined to detect collision based on this difference from a pre-defined set of conditions. This parameter captures the substantial change in speed during a collision thereby enabling the detection of accidents from its variation.\par

The Trajectory Anomaly ($\beta$) is determined from the angle of intersection of the trajectories of vehicles ($\theta$) upon meeting the overlapping condition C1. 

\begin{enumerate}
\item If $\theta \in (\theta_L$ $\theta_H)$, $\beta$ is determined from a pre-defined set of conditions on the value of $\theta$.
\item Else,  $\beta$  is determined from $\theta$ and the distance of the point of intersection of the trajectories from a pre-defined set of conditions.
\end{enumerate}

Thirdly, we introduce a new parameter that takes into account the abnormalities in the orientation of a vehicle during a collision. We determine this parameter by determining the angle ($\theta$) of a vehicle with respect to its own trajectories over a course of an interval of five frames. Since in an accident, a vehicle undergoes a degree of rotation with respect to an axis, the trajectories then act as the tangential vector with respect to the axis. By taking the change in angles of the trajectories of a vehicle, we can determine this degree of rotation and hence understand the extent to which the vehicle has underwent an orientation change. Based on this angle for each of the vehicles in question, we determine the Change in Angle Anomaly ($\gamma$) based on a pre-defined set of conditions.

Lastly, we combine all the individually determined anomaly with the help of a function to determine whether or not an accident has occurred. This function ${f(\alpha, \beta, \gamma)}$ takes into account the weightages of each of the individual thresholds based on their values and generates a score between ${0}$ and ${1}$. A score which is greater than ${0.5}$ is considered as a vehicular accident else it is discarded. This is the key principle for detecting an accident.

\section{Experimental evaluation}
\label{section3}
All the experiments were conducted on Intel(R) Xeon(R) CPU @ 2.30GHz with NVIDIA Tesla K80 GPU, 12GB VRAM, and 12GB Main Memory (RAM). All programs were written in ${Python - 3.5}$ and utilized ${Keras - 2.2.4}$ and ${Tensorflow - 1.12.0}$. Video processing was done using ${OpenCV 4.0}$.

\subsection{Dataset Used}

This work is evaluated on vehicular collision footage from different geographical regions, compiled from YouTube. The surveillance videos at 30 frames per second (FPS) are considered. The video clips are trimmed down to approximately 20 seconds to include the frames with accidents. All the data samples that are tested by this model are CCTV videos recorded at road intersections from different parts of the world. The dataset includes accidents in various ambient conditions such as harsh sunlight, daylight hours, snow and night hours. A sample of the dataset is illustrated in Figure \ref{fig:conditions}.

\subsection{Results, Statistics and Comparison with Existing models}

\begin{table}[h!]
\centering
\caption{Comparisons among the performance of other accident detection algorithms}
\label{table:comparisons}
\begin{tabular}{|l|c|c|c|}
\hline
Approach & Diff. Cameras & DR \% & FAR \% \\ \hline
\begin{tabular}[c]{@{}l@{}}Vision based \\ model (ARRS) \cite{ref15} \end{tabular} & 1 & 50 & 0.004 \\ \hline
\begin{tabular}[c]{@{}l@{}}Deep spatio-temporal \\ Model \cite{ref13}\end{tabular} & 7 & 77.5 & 22.5 \\ \hline
Proposed Framework & 45 & 71 & 0.53 \\ \hline
\end{tabular}
\end{table}

We estimate the collision between two vehicles and visually represent the collision region of interest in the frame with a circle as show in Figure \ref{fig:accidents}. We can observe that each car is encompassed by its bounding boxes and a mask. The magenta line protruding from a vehicle depicts its trajectory along the direction. In the event of a collision, a circle encompasses the vehicles that collided is shown.\par

The existing video-based accident detection approaches use limited number of surveillance cameras compared to the dataset in this work. Hence, a more realistic data is considered and evaluated in this work compared to the existing literature as given in Table  \ref{table:comparisons}. 

\begin{equation}\label{eq:DR}
 \text{Detection Ratio} = \frac{\text{Detected accident cases}}{\text{Total accident cases in the dataset}} \times 100 
\end{equation}
\begin{equation}\label{eq:FAR}
\text{False Alarm Rate} = \frac{\text{Patterns where false alarm occurs}}{\text{Total number of patterns}} \times 100
\end{equation}

The proposed framework achieved a detection rate of 71 \% calculated using Eq. \ref{eq:DR} and a false alarm rate of 0.53 \% calculated using Eq. \ref{eq:FAR}. The efficacy of the proposed approach is due to consideration of the diverse factors that could result in a collision.\par

\section{Conclusion and Future Works}
\label{section4}

In this paper, a new framework to detect vehicular collisions is proposed. This framework is based on local features such as trajectory intersection, velocity calculation and their anomalies. All the experiments conducted in relation to this framework validate the potency and efficiency of the proposition and thereby authenticates the fact that the framework can render timely, valuable information to the concerned authorities. The incorporation of multiple parameters to evaluate the possibility of an accident amplifies the reliability of our system. Since we are focusing on a particular region of interest around the detected, masked vehicles, we could localize the accident events. The proposed framework is able to detect accidents correctly with 71\% Detection Rate with 0.53\% False Alarm Rate on the accident videos obtained under various ambient conditions such as daylight, night and snow. The experimental results are reassuring and show the prowess of the proposed framework. However, one of the limitation of this work is its ineffectiveness for high density traffic due to inaccuracies in vehicle detection and tracking, that will be addressed in future work.  In addition, large obstacles obstructing the field of view of the cameras may affect the tracking of vehicles and in turn the collision detection.

\section{Acknowledgements}
We thank Google Colaboratory for providing the necessary GPU hardware for conducting the experiments and YouTube for availing the videos used in this dataset.

\bibliographystyle{IEEEtran}
\bibliography{ref}
\end{document}